\pdfoutput=1

\documentclass[conference]{IEEEtran}
\IEEEoverridecommandlockouts
\usepackage{cite}
\usepackage{amsmath,amssymb,amsfonts}
\usepackage{algorithmic}
\usepackage{graphicx}
\usepackage{textcomp}
\usepackage{xcolor}
\def\BibTeX{{\rm B\kern-.05em{\sc i\kern-.025em b}\kern-.08em
    T\kern-.1667em\lower.7ex\hbox{E}\kern-.125emX}}

\usepackage{hyperref}
\usepackage{xspace}
\usepackage{booktabs}
\usepackage{multirow}
\usepackage{cuted}
\usepackage{capt-of}

\newcommand{\method}{ChromaGS\xspace}

\newcommand{\fullmethod}{Text-Driven Semantic Editing of 4D Gaussian Avatars}

\makeatletter
\let\@oldmaketitle\@maketitle
\renewcommand{\@maketitle}{\@oldmaketitle
\begin{center}
    \includegraphics[width=\textwidth]{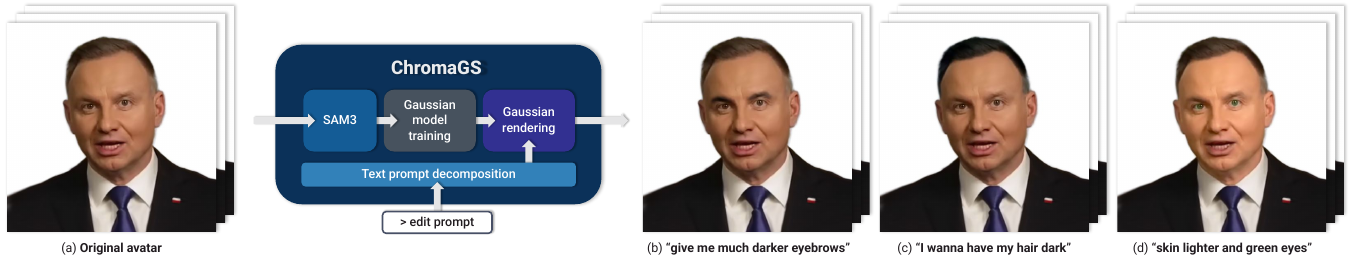}
    \captionof{figure}{\method enables real-time, language-guided color editing of 3D Gaussian head avatars. Given a trained avatar (left), users can instantly modify semantic region colors through natural language instructions, with edits applied directly at rendering time, bypassing the need for per-edit optimization.}
    \label{fig:teaser}
\end{center}

}
\makeatother

\begin{document}

\title{\method: \fullmethod}


\author{
  \IEEEauthorblockN{Antonio Canela Jim\'enez\textsuperscript{1,2,*} \quad and \quad Jordi S\`anchez-Riera\textsuperscript{2}}
  \vspace{0.15cm}
  \IEEEauthorblockA{\textsuperscript{1}\textit{Universitat Polit\`ecnica de Catalunya}}
  \IEEEauthorblockA{\textsuperscript{2}\textit{Institut de Rob\`otica i Inform\`atica Industrial, CSIC-UPC}}
  \IEEEauthorblockA{\textsuperscript{*}\textit{Corresponding author}}
}

\maketitle

\begin{abstract}
We present \method, a method for real-time, language-guided color editing of animatable 3D Gaussian head avatars.
Given a trained animatable avatar, users can instantly modify the color of semantic regions through natural language, with edits applied at render time and no retraining required.
Our key insight is to augment each Gaussian primitive with learned soft assignments to semantic regions and decompose colors into region-level base colors and Gaussian-level residuals.
This decomposition enables coherent color transfer: modifying a region's base color propagates naturally through all associated Gaussians while preserving fine appearance details encoded in residuals.
A two-stage language pipeline translates text instructions into target colors, supporting both absolute specifications and relative adjustments.
Unlike generative editing methods that may introduce unintended modifications, our approach provides deterministic, precisely localized semantic control.
Experiments demonstrate faithful appearance preservation and intuitive interaction across diverse subjects. Project page and code are available at: \url{https://a-canela.github.io/chromags/}

\end{abstract}


\section{Introduction}
\label{sec:introduction}
\begin{figure*}[t]
    \centering
    \includegraphics[width=\textwidth]{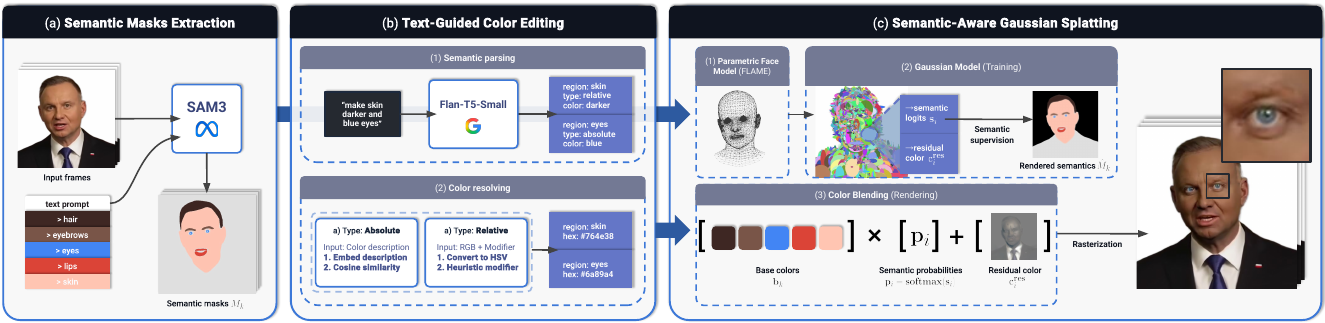}
    \caption{\textbf{(a)}~SAM3 extracts semantic masks from input frames. \textbf{(b)}~Text-guided editing: Flan-T5-Small parses instructions into structured text; a resolver handles absolute colors (palette lookup via embedding similarity) and relative modifiers (HSV shifts). \textbf{(c)}~Gaussians learn semantic logits (supervised by mask cross-entropy) and decompose color into base colors $\mathbf{b}_k$ plus residuals; editing replaces $\mathbf{b}_k$ while preserving fine details in residuals.}
    \label{fig:method}
\end{figure*}

Photorealistic 3D head avatars have become essential for applications ranging from telepresence and virtual reality to film production and gaming.
Recent advances in neural radiance fields~\cite{mildenhall2020nerf} and 3D Gaussian splatting~\cite{kerbl3Dgaussians} have enabled high-fidelity avatar reconstruction from monocular video, with animatable variants supporting real-time expression-driven rendering~\cite{qian2024gaussianavatars,Li_2025_CVPR}.

However, editing the appearance of these avatars remains challenging.
Users often want to modify attributes-eye color, skin tone, hair color-for creative customization or virtual try-on, without altering identity.
Existing generative editing methods based on diffusion models, while powerful, often introduce unintended modifications, struggle with precise localization, and cannot guarantee deterministic outputs~\cite{liu2025gaussianavatar}.

We present \method, enabling \emph{real-time, language-guided semantic color editing} of 3D Gaussian head avatars (Fig.~\ref{fig:teaser}), where users specify modifications through natural language and see results instantly.

Our key idea is to augment Gaussian primitives with semantic awareness and decompose their colors in a way that enables coherent region-level editing while preserving fine details.
A language interface translates text instructions into target colors, supporting both absolute and relative modifications.

Our contributions are: (1)~a semantic Gaussian representation with learned soft region assignments enabling localized color editing; (2)~a color decomposition that separates base colors from residual details; and (3)~a two-stage language pipeline for robust, intuitive color specification.
Experiments demonstrate instant, deterministic color editing with faithful appearance preservation across diverse subjects.

\section{Related Work}
\label{sec:related}
\noindent\textbf{Text-Guided 3D Gaussian Editing.}
Several methods address text-guided editing of 3D Gaussian scenes.
GaussianEditor~\cite{chen2024gaussianeditor}, Instruct-GS2GS~\cite{igs2gs}, and 3DSceneEditor~\cite{Yan_2026_WACV} leverage diffusion priors or open-vocabulary grounding for general scene editing, while LangSplat~\cite{qin2024langsplat} embeds CLIP features into Gaussians for open-vocabulary queries.
For head avatars, HeadStudio~\cite{zhou2024headstudio}, HeadArtist~\cite{liu2024headartist}, and HeadSculpt~\cite{han2023headsculpt} generate stylized heads from text via score distillation, and GaussianAvatar-Editor~\cite{liu2025gaussianavatar} and FaceG2E~\cite{wu2024text} apply diffusion-based optimization for text-guided appearance changes.
These generative approaches require per-edit optimization, may introduce unintended modifications, and cannot guarantee deterministic outputs.
In contrast, our method enables instant, deterministic color editing at render time without retraining.

\noindent\textbf{Palette-Based 3D Recoloring.}
A complementary line of work decomposes 3D scene appearance into a small set of palette colors that can be interactively edited.
PaletteNeRF~\cite{kuang2023palettenerf} and RecolorNeRF~\cite{gong2023recolornerf} introduced this idea for NeRF, representing each point's color as a linear combination of learnable palette bases.
PaletteGaussian~\cite{ren2024palettegaussian} extends the palette paradigm to 3D Gaussian Splatting, supporting manual, image-driven, and text-driven palette edits.
While effective for global and region-level recoloring of static scenes, these methods do not model animation, expression-dependent appearance, or semantic regions of a face, all of which are central to our setting.

\noindent\textbf{Semantic Editing in 3D Gaussian Scenes.}
Gaussian Grouping~\cite{ye2024gaussiangrouping} and OpenGaussian~\cite{wu2024opengaussian} assign semantic labels to Gaussians for segmentation-based scene editing.
However, these methods target general static scenes and do not handle animatable avatars or language-driven color specification.
Our approach combines learned semantic assignments with a color decomposition tailored for animatable head avatars, enabling region-specific editing that remains consistent across expressions and viewpoints.

\noindent\textbf{Semantic Segmentation and Language-Guided Editing.}
Vision-language models like SAM~\cite{kirillov2023segany} and SAM3~\cite{carion2025sam3segmentconcepts} enable open-vocabulary segmentation via text prompts.
For text-driven manipulation, diffusion-based approaches~\cite{rombach2021highresolution,brooks2022instructpix2pix} enable powerful but stochastic editing.
We leverage SAM3 for semantic mask extraction and propose a deterministic language-guided color editing pipeline.

\section{Method}
\label{sec:method}

\subsection{Semantic Masks Extraction}
\label{sec:masks}

For each semantic class $k$ and each training frame, we use SAM3~\cite{carion2025sam3segmentconcepts} with text prompts to generate binary masks $M_k$.
Iris masks are obtained from eye masks by detecting ellipses and warping an annular-shaped template.
Masks are made mutually exclusive by prioritizing smaller regions (e.g., iris over eye sclera) to ensure proper layering.
SAM3 exhibits strong temporal consistency across frames due to its concept-level understanding; moreover, since our optimization aggregates supervision from hundreds of frames, occasional segmentation noise is naturally absorbed during training.

\subsection{Text-Guided Color Editing}
\label{sec:language_model}

To enable intuitive color specification, we propose a two-stage architecture that decouples semantic parsing from color resolution, improving robustness by assigning simpler, well-defined tasks to each component.

\noindent\textbf{Semantic Parsing.}
We fine-tune Flan-T5-Small~\cite{2020t5} to parse natural language instructions into a structured intermediate format.
Given input text (e.g., ``red matte lipstick''), the model outputs a structured string: \emph{region, color, type} (e.g., \emph{lips, red matte, absolute}).
Multiple edits are separated by semicolons.
This simplified output space (region names, color descriptors, and edit types) is significantly easier to learn than direct RGB prediction.

\noindent\textbf{Color Resolving.}
The structured output is processed by a deterministic color resolver that handles two types of instructions:

\noindent\emph{Absolute.}
For absolute color instructions (e.g., ``olive skin''), we encode the color description using a pretrained sentence encoder (all-MiniLM-L6-v2)~\cite{reimers-2019-sentence-bert,wang2020minilm} and find the closest match via cosine similarity against pre-computed embeddings of region-specific palettes.
These palettes contain anatomically plausible colors (e.g., Fitzpatrick scale~\cite{fitzpatrick1988validity} tones for skin, natural iris colors based on melanin distributions) and serve as the target color space for each region.

\noindent\emph{Relative.}
For relative instructions (e.g., ``darker'', ``more saturated''), we convert the current region color to HSV and apply deterministic modifiers (e.g., reducing V for ``darker'', increasing S for ``more saturated'').

\subsection{Semantic-Aware Gaussian Splatting}
\label{sec:semantic_gaussians}

\noindent\textbf{Parametric Face Model.}
We build upon animatable 3D Gaussian representations~\cite{Li_2025_CVPR} where Gaussians are bound to FLAME~\cite{FLAME:SiggraphAsia2017}, enabling expression-driven animation together with linear blend skinning.

\noindent\textbf{Gaussian Model.}
Each Gaussian primitive $g_i$ is characterized by position $\boldsymbol{\mu}_i$, scaling $\boldsymbol{\sigma}_i$, rotation quaternion $\mathbf{q}_i$, opacity $\alpha_i$, and spherical harmonic color coefficients $\mathbf{c}_i$.
To enable localized color editing, we augment each Gaussian with \emph{semantic logits} $\mathbf{s}_i \in \mathbb{R}^K$, where $K$ is the number of predefined semantic classes.
Soft class probabilities are obtained via softmax: $\mathbf{p}_i = \text{softmax}(\mathbf{s}_i)$.

We supervise semantic assignments using the 2D masks from Sec.~\ref{sec:masks}.
During training, we render semantic probabilities using differentiable Gaussian rasterization:
\begin{equation}
    \hat{M}_k(\mathbf{u}) = \sum_{i} p_{i,k} \cdot \alpha_i \cdot G_i(\mathbf{u}),
\end{equation}
where $G_i(\mathbf{u})$ is the 2D Gaussian evaluated at pixel $\mathbf{u}$, and $p_{i,k}$ is the probability of Gaussian $i$ belonging to class $k$.
We apply a pixel-wise cross-entropy loss against blurred target masks:
\begin{equation}
    \mathcal{L}_{\text{sem}} = -\frac{1}{|\Omega|} \sum_{\mathbf{u} \in \Omega} \sum_k \tilde{M}_k(\mathbf{u}) \log \hat{M}_k(\mathbf{u}),
\end{equation}
where $\Omega$ is the foreground region and $\tilde{M}_k$ are soft targets obtained by applying per-class Gaussian blur (with class-specific $\sigma_k$) to the binary masks and renormalizing across classes.
This softens the supervision at region boundaries, where mask annotations are inherently imprecise, and provides smoother gradients than hard labels.

\noindent\textbf{Color Decomposition.}
We decompose Gaussian colors into two components: (1)~class-level \emph{base colors} $\mathbf{b}_k$, computed as the mean color from pixels within class $k$ masks across all training frames; and (2)~per-Gaussian \emph{residuals} $\mathbf{c}_i^{\text{res}}$ encoding local details such as texture, shading, and specular highlights.
The residuals are represented as learnable spherical harmonic coefficients that are combined with expression-dependent blendshapes via linear blend skinning~\cite{Li_2025_CVPR}: $\mathbf{c}_i^{\text{res}} = \mathbf{c}_i^{0} + \sum_l w_l \, \mathbf{c}_i^{l}$, where $\mathbf{c}_i^{0}$ is an expression-independent base residual, $\{\mathbf{c}_i^{l}\}$ are learned blendshape bases, and $w_l$ are projected FLAME expression coefficients.
This allows appearance details to vary naturally across facial expressions.

The final rendered color combines base colors, semantic probabilities, and residuals:
\begin{equation}
    \mathbf{c}_i^{\text{final}} = \sum_k p_{i,k} \cdot \mathbf{b}_k + \mathbf{c}_i^{\text{res}},
\end{equation}
where $p_{i,k} = \text{softmax}(\mathbf{s}_i)_k$ are the semantic probabilities.

To edit a region at inference, we replace its base color with a target value:
\begin{equation}
    \mathbf{c}_i^{\text{edited}} = \mathbf{c}_i^{\text{final}} + \sum_k p_{i,k} \cdot (\mathbf{b}_k^{\text{target}} - \mathbf{b}_k),
\end{equation}
where $\mathbf{b}_k^{\text{target}} = \mathbf{b}_k$ for unedited regions, so only modified regions contribute to the color shift while residuals preserve local details.
The soft semantic assignments ensure smooth transitions at region boundaries.

\noindent\textbf{Cross-Region Adaptation.}
Each semantic class is a weighted combination of mask channels, where the weights act as user-exposed parameters, providing a straightforward yet effective mechanism to encode prior knowledge about anatomical color correlations between regions.
For instance, the \emph{skin} class is driven mainly by the skin mask but also receives minor contributions from lips ($0.3$) and eyebrows ($0.2$), so that editing skin tone shifts pigment-correlated regions coherently rather than producing an unnatural disconnect at their boundaries.

\subsection{Training Objective}
\label{sec:training}

The full training objective combines photometric reconstruction with semantic supervision:
\begin{equation}
    \mathcal{L} = \lambda_{\text{rgb}} \mathcal{L}_{\text{rgb}} + \lambda_{\text{sem}} \mathcal{L}_{\text{sem}},
\end{equation}
where $\mathcal{L}_{\text{rgb}}$ is the standard L1 photometric loss and $\mathcal{L}_{\text{sem}}$ is the cross-entropy semantic loss.
We use $\lambda_{\text{rgb}} = 1.0$ and $\lambda_{\text{sem}} = 0.1$ in all experiments.

\section{Experiments}
\label{sec:results}
\noindent\textbf{Implementation Details.}
We implement \method on top of an animatable Gaussian splatting framework~\cite{Li_2025_CVPR}.
We define $K=6$ semantic classes (hair, eyebrows, eyes, lips, skin, plus background).
Training runs for 50k iterations on a single NVIDIA~RTX~5090, taking approximately 15~minutes.

\noindent\textbf{Datasets and Evaluation Setup.}
We evaluate on scenes from FaceScape~\cite{zhu2023facescape} and INSTA~\cite{INSTA:CVPR2023}, featuring diverse skin tones, eye colors, and hair styles.
Each scene is edited with 14 diverse color edits spanning hair, eyes, skin, lips, and eyebrows, yielding approximately 10k edited frames per method at $512 \times 512$ resolution.

\begin{figure}[t]
    \centering
    \includegraphics[width=\columnwidth]{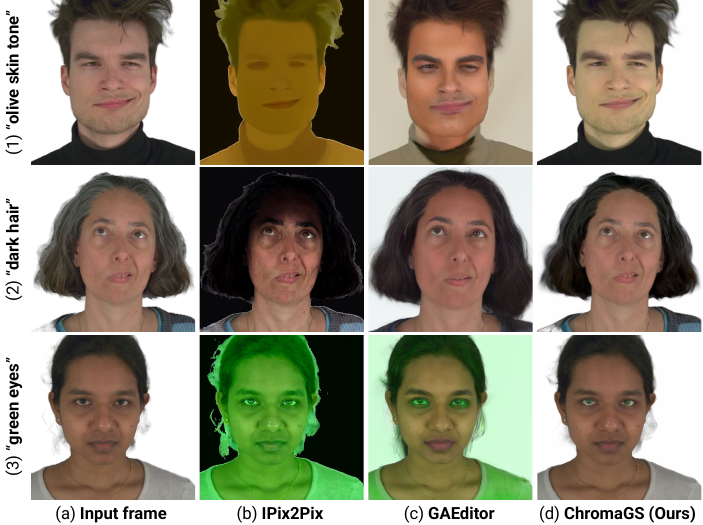}
    \caption{\textbf{Qualitative comparison.} Results for skin, hair, and eyes edits across three subjects. Our method produces localized, natural color changes that preserve identity and fine details (e.g., specular highlights, skin texture). Both GaussianAvatar-Editor~\cite{liu2025gaussianavatar} and InstructPix2Pix~\cite{brooks2022instructpix2pix} exhibit color bleeding into unrelated regions and over-saturated tones.}
    \label{fig:results}
\end{figure}

\noindent\textbf{Metrics.}
We measure: (1)~\textbf{Identity Preservation}$\uparrow$: ArcFace cosine similarity between edited and unedited renders; (2)~\textbf{CLIP Score}$\uparrow$: alignment between rendered image and target text prompt, averaged across three CLIP backbones (ViT-L/14, ViT-B/16, ViT-B/32); (3)~\textbf{Context Preservation}$\uparrow$: computed as $1 - \text{LPIPS}$ outside the target region, measuring how well non-target areas are preserved after editing.

\subsection{Comparison with Baselines}

\begin{table}[t]
    \centering
    \caption{\textbf{Runtime comparison.} Our method requires a one-time $\sim$800\,s training, after which any subsequent edit is applied instantly ($\sim$50\,ms). In contrast, GaussianAvatar-Editor~\cite{liu2025gaussianavatar} relies on a coupled training-and-editing loop, requiring a full $\sim$800\,s optimization for every new edit. InstructPix2Pix~\cite{brooks2022instructpix2pix} requires no per-subject optimization as it processes each frame independently via diffusion.}
    \label{tab:timing}
    \small
    \begin{tabular*}{\columnwidth}{@{\extracolsep{\fill}}lccc@{}}
        \toprule
        Method & Train$\downarrow$ & Edit$\downarrow$ & Render$\uparrow$ \\
        \midrule
        IPix2Pix~\cite{brooks2022instructpix2pix} & N/A & $\sim$400 ms & N/A \\
        GAEditor~\cite{liu2025gaussianavatar} & \multicolumn{2}{c}{\quad \leftarrowfill \quad $\sim$800 s \quad \rightarrowfill \quad} & $>$100 fps \\
        \method (Ours) & $\sim$800 s & $\sim$\textbf{50 ms} & $>$100 fps \\
        \bottomrule
    \end{tabular*}
\end{table}

\begin{table}[t]
    \centering
    \caption{\textbf{Quantitative comparison with baselines.} Our method achieves significantly better identity and context preservation. The baselines' higher CLIP scores reflect a limitation of the metric itself: CLIP measures global image-text alignment and thus rewards visually dominant color shifts, even when they bleed into unintended regions.}
    \label{tab:comparison}
    \small
    \begin{tabular*}{\columnwidth}{@{\extracolsep{\fill}}lccc@{}}
        \toprule
        Method & CLIP$\uparrow$ & Identity$\uparrow$ & Context$\uparrow$ \\
        \midrule
        IPix2Pix~\cite{brooks2022instructpix2pix} & \textbf{0.231} & 0.691 & 0.790 \\
        GAEditor~\cite{liu2025gaussianavatar} & 0.229 & 0.863 & 0.790 \\
        \method (Ours) & 0.212 & \textbf{0.986} & \textbf{0.980} \\
        \bottomrule
    \end{tabular*}
\end{table}

We compare against \textbf{GaussianAvatar-Editor}~\cite{liu2025gaussianavatar}, a text-guided Gaussian head avatar editing method using diffusion priors, and \textbf{InstructPix2Pix}~\cite{brooks2022instructpix2pix}, an image-conditioned diffusion model that applies text-guided edits frame-by-frame.
While other text-guided 3D editing methods exist, they target static scenes rather than animatable avatars~\cite{chen2024gaussianeditor}, \cite{ye2024gaussiangrouping}, \cite{igs2gs}, \cite{Yan_2026_WACV}, \cite{ren2024palettegaussian} or edit generated heads instead of reconstructions from real video~\cite{zhou2024headstudio}, \cite{liu2024headartist}, \cite{han2023headsculpt}, \cite{wu2024text}, making a direct comparison neither applicable nor fair; the selected baselines represent the closest available approaches to our setting.
Fig.~\ref{fig:results} shows color editing results across multiple subjects and semantic regions. Our method produces consistent, localized color changes that preserve identity and fine details (e.g., specular highlights, texture variations).
Edits propagate naturally across different expressions and viewpoints due to the 3D Gaussian representation.

Table~\ref{tab:comparison} presents quantitative comparisons using the pre-trained scenes publicly released by GaussianAvatar-Editor~\cite{liu2025gaussianavatar}.
Our method achieves substantially higher identity and context preservation scores, reflecting its ability to confine edits precisely to the target region.
The baselines obtain slightly higher CLIP scores despite producing visibly less accurate edits (Fig.~\ref{fig:results}).
This apparent paradox arises because CLIP measures holistic image-text similarity: a diffusion-based method that tints the entire face green when asked for ``green eyes'' produces a stronger global color signal that CLIP rewards, even though the edit is spatially incorrect.
Our method, by design, restricts color changes to the semantically correct region, yielding a subtler global shift that CLIP is less sensitive to, a trade-off we consider desirable.

Beyond metric performance, the methods differ substantially in practical usability (Table~\ref{tab:timing}).
GaussianAvatar-Editor couples editing and training into a single optimization loop of $\sim$800~s, which must be repeated for every new edit.
Our method shares a comparable one-time training cost, but decouples editing entirely: once trained, any color change is applied in $\sim$50~ms (which mainly corresponds to the language model inference mapping the text prompt to color parameters), three orders of magnitude faster, enabling interactive exploration of diverse edits.
InstructPix2Pix operates frame-by-frame with no 3D awareness, which not only limits throughput but also introduces severe temporal flickering across the sequence.

\subsection{Ablation Studies}

\begin{figure}[t]
    \centering
    \includegraphics[width=\columnwidth]{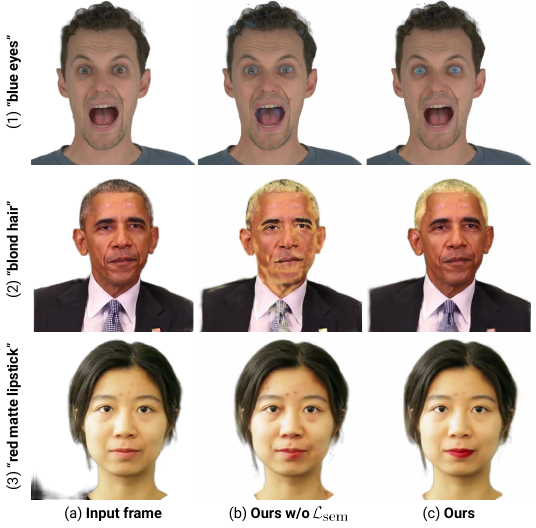}
    \caption{\textbf{Ablation: semantic supervision.} Without $\mathcal{L}_{\text{sem}}$, edits spread to unrelated areas: blue eyes tint the hair and tongue, blonde hair bleeds into the face and clothing, and red lipstick affects the skin color as well. Our full model produces clean, confined edits.}
    \label{fig:ablation}
\end{figure}

\noindent\textbf{Semantic Supervision.}
Table~\ref{tab:ablation_semantic} and Fig.~\ref{fig:ablation} show that omitting semantic supervision ($\lambda_{\text{sem}}=0$) prevents Gaussians from learning distinct region assignments, causing edits to erroneously affect unrelated facial regions. This confirms $\mathcal{L}_{sem}$ is essential for regularizing primitives into coherent semantic clusters, effectively anchoring color modulation to the intended targets through robust semantic disentanglement.

\begin{table}[t]
    \centering
    \caption{\textbf{Quantitative ablation: Semantic supervision.} Without $\mathcal{L}_{\text{sem}}$, color changes spread beyond the target region, degrading all three metrics. Notably, CLIP also drops: unlike the diffusion baselines that apply intense, globally visible color shifts, the unguided leakage here dilutes the edit rather than amplifying it.}
    \label{tab:ablation_semantic}
    \small
    \begin{tabular*}{\columnwidth}{@{\extracolsep{\fill}}lccc@{}}
        \toprule
        Method & CLIP$\uparrow$ & Identity$\uparrow$ & Context$\uparrow$ \\
        \midrule
        Ours w/o $\mathcal{L}_{\text{sem}}$ & 0.198 & 0.969          & 0.967          \\
        Ours                                & \textbf{0.213} & \textbf{0.990} & \textbf{0.986} \\
        \bottomrule
    \end{tabular*}
\end{table}

\section{Limitations and Future Work}
\label{sec:limitations}
Localization quality depends on the underlying semantic segmentation~\cite{carion2025sam3segmentconcepts}. Our approach enables uniform shifts but precludes texture/material modifications (e.g., metallic skin, freckles) and geometric edits (e.g., reshaping features). High-frequency structures like hair may lack photorealism, as residual-based decomposition struggles with such intricate details. Additionally, color resolution is constrained by curated palettes; out-of-palette tones default to the nearest entry, potentially misaligning with user intent. Finally, reconstruction quality is bounded by the accuracy of upstream FLAME tracking and the base Gaussian avatar model.

Future work will explore spatially-varying material editing, learning richer appearance priors from multi-identity datasets, expanding the semantic vocabulary toward open-set region discovery, and developing palette-free color strategies.

\section{Conclusion}
\label{sec:conclusion}
We introduced \method for real-time, language-driven color editing of animatable 3D Gaussian avatars. By coupling learned segmentation with color decomposition, our framework enables localized, deterministic control without requiring per-edit optimization. Our two-stage pipeline effectively bridges natural language with precise absolute and relative color adjustments. Ultimately, replacing diffusion-based optimization with a single forward pass allows \method to make high-fidelity avatar customization practical for interactive gaming and virtual try-on.

\section{Acknowledgments}
\label{sec:acknowledgments}
This work has been partially supported by project {\small PID2023-151184OB-I00}, funded by the Ministerio de Ciencia e Innovación (MCIN) / Agencia Estatal de Investigación (AEI) / 10.13039/501100011033, by the European Social Fund (ESF) ``Investing in your future'' and by \textit{ERDF, EU}; and by the industrial doctorate 2021 DI 00102 funded by the Government of Catalonia.

\bibliographystyle{IEEEtran}
\bibliography{references}

\end{document}